\documentclass[10pt,twocolumn,letterpaper]{article}

\usepackage{cvpr}              %

\usepackage{microtype}

\usepackage[accsupp]{axessibility}  %

\usepackage{multirow}
\usepackage{tabularx}

\definecolor{cvprblue}{rgb}{0.21,0.49,0.74}
\usepackage[pagebackref,breaklinks,colorlinks,allcolors=cvprblue]{hyperref}

\def\paperID{} %
\def\confName{CVPR}
\def\confYear{2026}

\title{VENI: Variational Encoder for Natural Illumination}

\author{
    Paul Walker$^{1}$ \enspace
    James A. D. Gardner$^{2,3}$ \enspace
    Andreea Ardelean$^{1}$ \enspace
    William A. P. Smith$^{2,3}$ \enspace
    Bernhard Egger$^1$ \vspace{0.2cm}\\
    $^1$Friedrich-Alexander-Universität Erlangen-Nürnberg\\
    $^2$University of York \quad $^3$pxld.ai\\
    \small \url{https://paul-pw.github.io/veni} \vspace{-6mm}
}

\begin{document}
\twocolumn[{%
\renewcommand\twocolumn[1][]{#1}%
\maketitle \thispagestyle{empty}
\begin{center}
    \centering
\resizebox{\textwidth}{!}{
    \includegraphics[width=\textwidth]{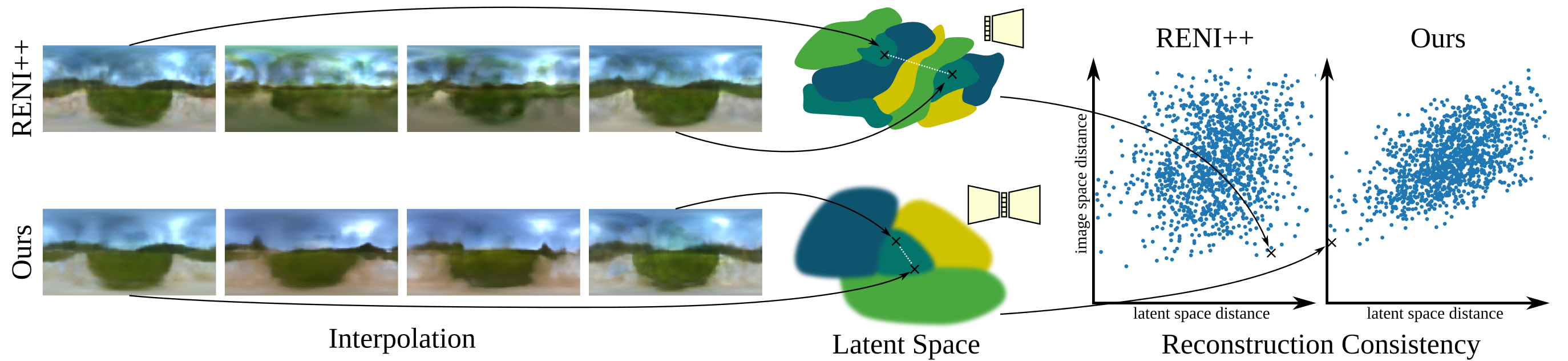}
}
    \captionof{figure}{We build a rotation-equivariant variational autoencoder model to address the limitations of the current state-of-the-art illumination prior (RENI++). Our model produces a better-behaved latent space, which we demonstrate by evaluating its uniqueness through optimization of random latent codes and its reconstruction consistency via the correlation between latent space and image space distances.}
    \label{fig:teaser}
\end{center}
}]
\begin{abstract}

Inverse rendering is an ill-posed problem, but priors such as illumination priors can help simplify it.
Existing work either disregards the spherical and rotation-equivariant nature of illumination environments or does not provide a well-behaved latent space.
We propose a rotation-equivariant variational autoencoder that models natural illumination on the sphere without relying on 2D projections.
To preserve the SO(2)-equivariance of environment maps, we use a novel Vector Neuron Vision Transformer (VN-ViT) as encoder and a rotation-equivariant conditional neural field as decoder.
In the encoder, we reduce the equivariance from SO(3) to SO(2) using a novel SO(2)-equivariant fully connected layer, an extension of Vector Neurons.
We show that our SO(2)-equivariant fully connected layer outperforms standard Vector Neurons when used in our SO(2)-equivariant model.
Compared to previous methods, our variational autoencoder enables smoother interpolation in latent space and offers a more well-behaved latent space.

\end{abstract}
    
\section{Introduction}
\label{sec:intro}

Inverse rendering is ill-posed, an image can be created by multiple combinations of shape, material and lighting~\cite{basrelief1999, inverseNMF2024}. To solve this, humans draw on strong, learned priors over the space of possible illuminations~\cite{visualperception2019}, especially a lighting-from-above prior~\cite{visualperception2019, lightabove2009, lightabove2010}, and it has been shown that humans perform better on inverse rendering tasks under natural illumination conditions, with performance degrading under artificial illumination~\cite{illuminationperception2003}.
Whilst natural illumination is complex and challenging to model, it displays statistical regularities~\cite{realworldillumination2004}, particularly in outdoor scenes. For example, the strongest sources of illumination are the sun and the skylight, which produce only a limited range of colors. 
Additionally, illumination environments have a canonical up direction, with any rotation about the vertical direction being equally likely.

However, instead of relying on learned priors for illumination, it is still common in inverse rendering to reconstruct the lighting directly~\cite{inverseshlambertian2003, indirectinverse2022, inversescene2021, wang2025materialist}
or to rely on simple statistical models of spherical harmonics parameters of custom datasets~\cite{illuminationprior2015, faceilluminationprior2018, naturalilluminationprior2022, inversescene2013}.

Recently, there has been some work to build natural illumination priors using spherical neural fields~\cite{NeuralPreIntegratedLighting2021, RENI++2023, RENI2022}. Such continuous representations are useful in an inverse rendering setting \cite{skylimit2025,spar3d2025} since the incident illumination can be queried from any direction. 
Some of these models \cite{RENI++2023, RENI2022} exploit the vertical rotation-equivariance inherent to illumination environments, leading to a more efficient model and negating the need for rotation augmentation during training. However, they are constructed as decoder-only architectures due to the difficulty of building a rotation-equivariant encoder. For example, RENI++ \cite{RENI++2023} employs an autodecoder architecture~\cite{park2019deepsdf}, in which latent codes are randomly initialized for each training and testing image and jointly optimized with the model. This leads to two key limitations. First, similar images are often initialized with completely different latent codes, which can cause the model to represent the same image multiple times within its latent space, \ie, the latent space is not unique.
Second, the training cannot scale to large datasets since a latent code must be optimized per-image and the latent space uniqueness further degrades as large numbers of similar images are initialized with different latent codes.

To overcome these two limitations while retaining rotation-equivariance and the benefits of a neural field representation, we propose a novel rotation-equivariant variational autoencoder architecture for spherical signals. Following the principles of the Vector Neuron~\cite{VectorNeurons2021} framework, we design a novel vision transformer \emph{encoder} that computes the latent codes, which are then mapped back to the image space via a neural field decoder.
This enables us to use much larger datasets, as we no longer need to explicitly optimize a latent code for each image, but instead encode the image to a latent code via a forward pass through our rotation-equivariant encoder.
By construction, encoding two similar images now leads to two similar latent codes, thereby addressing the weakness of RENI++ and improving the uniqueness of the latent space.
This enables smooth illumination transitions through latent space interpolation.

To summarize, our contributions are:
\begin{itemize}
    \item A rotation-equivariant panoramic vision transformer encoder enabled by a novel SO(2)-equivariant extension to Vector Neurons
    \item A rotation-equivariant natural illumination prior with a well-behaved, unique latent space
    \item Extensive evaluation of latent space uniqueness and reconstruction consistency using intuitive metrics
\end{itemize}

\section{Related Work}
\label{sec:related-work}

\noindent\textbf{Lighting Representations.}
The lighting at any point in 3D space can be represented as a spherical signal
that defines the incident light intensity and color from every direction.
In computer graphics, it is often assumed that the illumination environment is the same across all points in the scene, \ie, it models distant illumination.
The illumination environment is often modeled by an environment map~\cite{sphericalgaussians2006, sphericalHarmonics2001}, which is a projection of the spherical signal onto a 2D image via  equirectangular projection or cube mapping~\cite{cubemap1986}.
However, both projections introduce some level of distortion, resulting in irregular sampling with respect to the original spherical image.
Since available datasets are typically provided as equirectangular projected images~\cite{lavalindoor2017, RENI2022, streetlearn2019}, the associated sampling issues need to be addressed during training.

Alternative representations for illumination environments used in inverse rendering problems are Spherical Harmonics (SH)~\cite{sphericalHarmonics2001, naturalilluminationprior2022, faceilluminationprior2018, inverseshlambertian2003, inversescene2013} 
and Spherical Gaussians (SG)~\cite{sphericalgaussians2006, indirectinverse2022, InverseNerf2021, inversescene2021}.
Since both express illumination environments as spherical signals, they neither introduce distortion nor suffer from irregular sampling.
Still, Boss \etal \cite{NeuralPreIntegratedLighting2021} and Gardner \etal \cite{RENI++2023} have shown that neural field-based light modeling can capture more detailed environments using fewer parameters than SG or SH, so we follow their approach for our decoder.

\noindent\textbf{Illumination Priors.}
Many inverse rendering works do not rely on illumination priors and instead reconstruct lighting directly~\cite{inverseshlambertian2003, indirectinverse2022, inversescene2021, wang2025materialist}. This often leads to unrealistic lighting estimations and hinders the estimation of other parameters, such as albedo. Since real world illumination exhibits some regularities~\cite{realworldillumination2004}, illumination priors have been shown to help constrain the inverse rendering task ~\cite{skylimit2025}.

Recent works in light estimation focus on outpainting a 360° HDR environment map from an LDR crop to estimate the lighting in a scene~\cite{emlight2021, stylelight2022, everlight2023}.
This introduces an implicit prior on the illumination.
Zhan \etal \cite{emlight2021} and Dastjerdi \etal \cite{everlight2023} first predict the location of the light sources from the LDR crop using SG and then use a generative adversarial network (GAN) to reconstruct the full HDR image based on the LDR crop and the light source prediction. %
Wang \etal \cite{stylelight2022} train a two branch StyleGAN~\cite{karras2019style} to generate LDR and HDR panoramas. Using this, they propose a GAN inversion method to find the latent code of an LDR crop to predict the HDR panorama. 
Phongthawee \etal \cite{DiffusionLight2024} propose to inpaint a chrome ball into a scene to estimate the lighting instead of directly outpainting a 360° panorama. This can also be seen as an outpainting method, since the inpainted chrome ball shows what is behind the camera.
While these models can reconstruct high quality illumination environments, they require that a part of the illumination environment is visible in the scene. Without which, these models cannot be used as an illumination prior.  %

Earlier works that do not have this limitation proposed statistical models over SH parameters of custom datasets~\cite{illuminationprior2015, faceilluminationprior2018, naturalilluminationprior2022, inversescene2013},
while more recent works rely on neural approaches~\cite{faceilluminationprior2020, NeuralPreIntegratedLighting2021, RENI++2023}. 
Sztrajman \etal \cite{faceilluminationprior2020} build an environment map convolutional autoencoder
and Boss \etal \cite{NeuralPreIntegratedLighting2021} use a neural field to model pre-integrated lighting.
Closest to our approach is RENI++ \cite{RENI++2023}, which proposes a natural illumination prior that is rotation-equivariant by design. RENI++ uses an autodecoder architecture~\cite{park2019deepsdf}, meaning that latent codes are initialized randomly per training/testing image and then jointly optimized with the model~\cite{RENI++2023}.
This approach results in a latent space capable of representing an image by multiple, different latent codes, \ie, the latent space is not unique. This adversely affects downstream performance. Our autoencoder approach overcomes this limitation, leading to a more well-behaved latent space.

\noindent\textbf{Rotation-equivariance.}
Lighting environments are inherently rotation-equivariant. Any rotation around the up-axis results in another valid illumination environment.
This property has been largely overlooked in priors for natural illumination~\cite{emlight2021,everlight2023,stylelight2022,DiffusionLight2024, illuminationprior2015, faceilluminationprior2018}, while other works try to achieve it via data augmentation~\cite{faceilluminationprior2020, naturalilluminationprior2022}.
rotation-equivariance in general is an important property in computer vision \cite{GeometricDL2021}. %
Some methods design rotation-equivariant models using convolutions with steerable kernels~\cite{steerablecnns2017, tensorfield2018, 3dsteerablecnns2018, functionalSO32021}.
Another approach is proposed by SE(3) Transformer~\cite{se3transformer2020}, which introduces a rotation-equivariant attention mechanism by combining attention~\cite{AttentionNeed2017} with tensor field convolution~\cite{tensorfield2018}. %
Steerable kernels and SE(3) transformers are restricted to convolutions, which limits their applicability to non-convolutional models.

A more general framework for building SO(3)-equivariant neural networks is Vector Neurons~\cite{VectorNeurons2021}. By providing basic SO(3)-equivariant building blocks, such as linear and ReLU layers, it enables the design of a wide range of equivariant model architectures.
For example, VN-Transformer~\cite{VNTransformer2023} applies the Vector Neuron framework to transformer models. 
We rely on VN-Transformer to achieve rotation-equivariance in our encoder.
The Vector Neurons approach was also used by RENI and RENI++ to build a rotation-equivariant prior for natural illumination~\cite{RENI2022, RENI++2023}.
While RENI uses the full Gram matrix, to achieve rotation-equivariance at the cost of $O(n^2)$ complexity relative to the size of the latent space,
RENI++ uses the VN-Invariant layer~\cite{VectorNeurons2021}, reducing complexity to $O(n)$~\cite{RENI++2023}. We follow the approach of RENI++ to obtain a rotation-equivariant decoder.

\noindent\textbf{Panoramic Vision Transformers.}
360° panoramic images present inherent challenges, as the 2D projection of such data (\eg, equirectangular projection, cube mapping) inevitably introduces distortion. 
Based on the transformer architecture~\cite{AttentionNeed2017}, Vision Transformers use global self-attention to solve many computer vision tasks~\cite{VisionTransformer2021}, but standard Vision Transformers are not designed to handle distortion. %
Shen \etal \cite{PannoFormer2022} propose the use of sphere tangent-patches to remove the negative effects of distortion when using vision transformers with 360° equirectangular projected images, while other methods address it via 
deformable convolutions~\cite{PanoViT2022, PanoViT2023, Laformer2023}.
Yun \etal \cite{PanoViT2022} use deformable convolutions with fixed offsets as linear projection into the transformer encoder. %
Zhao \etal \cite{PanoViT2023} propose a distortion-aware transformer block that uses deformable convolution with learnable offsets.
Yuan \etal \cite{Laformer2023} use deformable convolutions with learnable offsets in the output projection of a 360° image segmentation model. %

All of these methods address the distortion introduced when mapping the spherical 360° image to 2D, but they disregard the true spherical 3D nature of  panoramic images.
Consequently, all of them must explicitly account for such distortions. In contrast, we operate directly in the 3D domain and build a vision transformer upon the VN-Transformer~\cite{VNTransformer2023}.
This approach eliminates the need for distortion handling, as no projection from 3D to 2D is involved.
The technique we use is closest to early fusion, as proposed in VN-Transformer~\cite{VNTransformer2023}.

\section{Methods}
\label{sec:methods}

Our goal is to learn a prior for natural illumination with a structured latent space, which will make operations in the latent space (\eg interpolations) more semantically meaningful.
Building on RENI++~\cite{RENI++2023}, we address their architectural issues by implementing a rotation-equivariant variational autoencoder that lets the model implicitly learn a structured latent space.
We adopt the RENI++ decoder and introduce a novel Vector Neuron Vision Transformer encoder.

\begin{figure}
  \centering
   \includegraphics[width=0.8\linewidth]{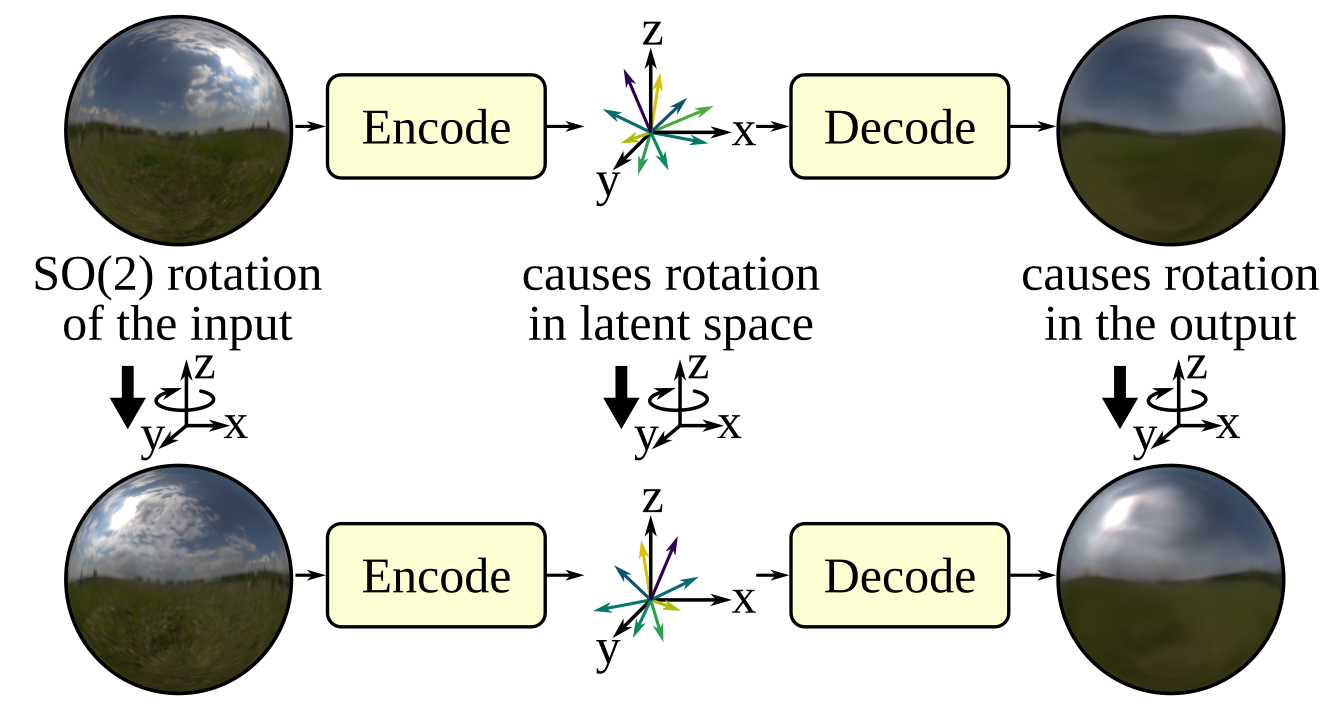} 

   \caption{Our model is rotation-equivariant, meaning that a rotation of the input environment map leads to a corresponding rotation in latent space and in the reconstructed environment map.}
   \label{fig:rotating_model}
   \vspace{-0.1in}
\end{figure}

\noindent\textbf{SO(2)-equivariance.}
A model is rotation-equivariant if a rotation of the input induces the same rotation of the output, as visualized in Figure~\ref{fig:rotating_model}.
For natural, outdoor illumination, where there is always a well-defined horizon line and up-axis,
only rotations around the up-axis lead to other realistic illumination environments. 
Rotations around other axes will result in unrealistic illumination environments. Therefore, we design an SO(2)-equivariant model.

\noindent\textbf{SO(2)-equivariant fully connected  layer.}
Vector Neurons, as proposed by Deng \etal in~\cite{VectorNeurons2021}, provide a framework 
for SO(3)-equivariant networks.
We deal with multidimensional data (\ie, direction and color), but only desire equivariance in 
the $x$ and $y$ dimensions (\ie, SO(2)-equivariance). Rotations around any other axis than the up-axis, particularly rotations of the color vectors, produce unrealistic illumination environments. Therefore, these dimensions should remain invariant to rotations in the $xy$-plane.
Since both the invariant dimensions and the equivariant dimensions contain important information, 
we aim for them to influence each other while preserving rotation-equivariance in the $x$ and $y$ dimensions and invariance in the other dimensions.

We extend Vector Neurons by combining invariant operations and equivariant operations in one neuron, resulting in SO(2)-equivariance.
Given the input, $\mathbf{X}=\left[\mathbf{X}_\textit{eq},\mathbf{X}_\textit{inv}\right]\in\mathbb{R}^{d_\textit{in}\times (2+c_\textit{inv})}$, 
consisting of the equivariant component, $\mathbf{X}_\textit{eq}\in\mathbb{R}^{d_\textit{in}\times 2}$, in our case, the $x$ and $y$ dimensions
and the invariant component, $\mathbf{X}_\textit{inv}\in\mathbb{R}^{d_\textit{in}\times c_\textit{inv}}$,  where $c_\textit{in}=4$ is the number of invariant input dimensions, here, the $z$ and color dimensions, and $d_{in}$ is the number of inputs to the fully connected layer,
we arrive at the intermediate values:
\begin{align}
    \mathbf{T}_{\textit{inv}} &= \left[\mathbf{X}_\textit{inv},\mathbf{1}_{d_\textit{in}\times 1}\right]\in\mathbb{R}^{d_\textit{in}\times (c_\textit{inv}+1)} \\
    \mathbf{T}'_{\textit{inv}} &= \left[ \mathbf{X}_\textit{inv},\lVert \mathbf{X}_\textit{eq}\rVert\right]\in\mathbb{R}^{d_\textit{in}\times (c_\textit{inv}+1)}
\end{align}
where the concatenation with the column of ones in $\mathbf{T}_\textit{inv}$ allows us to directly incorporate a bias in the bilinear combination with the equivariant inputs $\mathbf{X}_\textit{eq}$. 
In $\mathbf{T}'_\textit{inv}$,  the concatenation with $\lVert\mathbf{X}_\textit{eq}\rVert$, which is the L2 norm in the last (equivariant) dimensions, allows the invariant outputs to depend on the equivariant inputs without losing invariance.
Given the weight and bias matrices,
\[
  \mathbf{W}_\textit{eq}\in\mathbb{R}^{d_\textit{out}\times(c_\textit{inv}+1)\times d_\textit{in}}\text{, } \mathbf{W}_\textit{inv}\in\mathbb{R}^{d_\textit{out}\times c_\textit{inv} \times d_\textit{in} \times (c_\textit{inv}+1)}
  \]
  \[\mathbf{B}_\textit{inv}\in\mathbb{R}^{d_\textit{out}\times c_\textit{inv}}\text{,}\]
we define an SO(2)-equivariant fully connected layer, $f\left(\mathbf{X}, \mathbf{W}_\textit{eq},\mathbf{W}_\textit{inv},\mathbf{B}_\textit{inv}\right)$, as follows:
\begin{align}
\mathbf{Y}_{\textit{eq},o,v} &= \sum_{i=1}^{d_\textit{in}}\sum_{k=1}^{c_\textit{inv}+1} \mathbf{W}_{\textit{eq},o,k,i}\cdot \mathbf{T}_{\textit{inv},i,k}\cdot \mathbf{X}_{\textit{eq},i,v} \\
  \mathbf{Y}_{\textit{inv},o,v} &=\sum_{i=1}^{d_\textit{in}}\sum_{k=1}^{c_\textit{inv}+1}\mathbf{W}_{\textit{inv},o,v,i,k}\cdot \mathbf{T}'_{\textit{inv},i,k}+\mathbf{B}_{\textit{inv},o,v}\\
  \mathbf{Y} &= f\left(\mathbf{X}, \mathbf{W}_\textit{eq},\mathbf{W}_\textit{inv},\mathbf{B}_\textit{inv}\right) = \left[ \mathbf{Y}_\textit{eq},\mathbf{Y}_\textit{inv}\right]
\end{align}
Where $\mathbf{Y}_\textit{eq}$ is a bilinear combination of the equivariant input $\mathbf{X}_\textit{eq}$ and the invariant input $\mathbf{X}_\textit{inv}$ and $\mathbf{Y}_\textit{inv}$ is a linear combination of the invariant input $\mathbf{X}_\textit{inv}$ and the length of the equivariant input vectors $\lVert\mathbf{X}_\textit{eq}\rVert$.
Proof of SO(2)-rotation-equivariance can be found in the supplementary material.

\begin{figure*}
  \centering
   \includegraphics[width=0.9\linewidth]{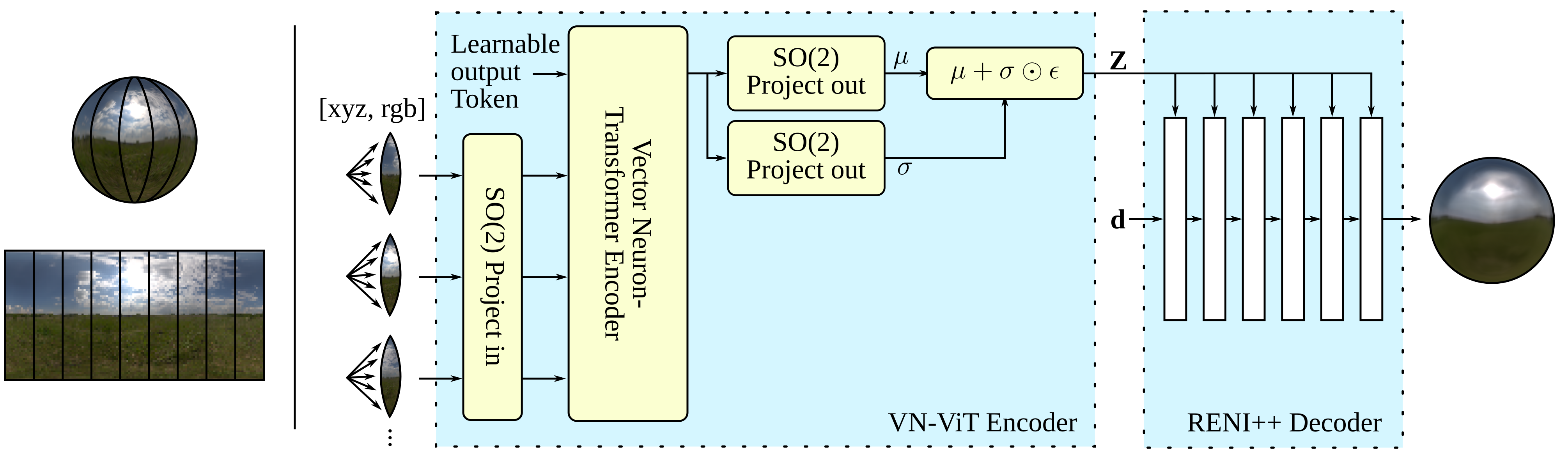} 

   \caption{Model Overview: We adapt the Vision Transformer architecture to SO(2) equivariance on spherical images. Splitting the 360° spherical image into vertical stripes and embedding the direction vectors and color values of the pixels in each patch using an SO(2)-equivariant projection. Then we feed the sequence into a Vector Neuron-Transformer. We use a class token that is learnable in non-equivariant dimensions and zero in equivariant dimensions. The output is projected to $\mu$ and $\log(\sigma^2)$ and reparameterized. A rotation of the input by a multiple of the patch width results in the output of the encoder being rotated in the same way. The output latent code of the encoder is fed into the RENI++ decoder.}
   \label{fig:model}
\end{figure*}

\noindent\textbf{Vector Neuron Vision Transformer (VN-ViT).}
Our key contribution is a Vision Transformer and Vector Neuron-based encoder (VN-ViT), that can encode spherical images to SO(2) rotation-equivariant latent codes.
Figure~\ref{fig:model} shows an overview of the full method.
The model architecture is based on vision transformers~\cite{VisionTransformer2021}, but since we work with spherical images and require rotation-equivariance, we replace all ViT components with their Vector Neuron counterparts~\cite{VectorNeurons2021}.
We cannot use positional encoding, since that would break equivariance.
Instead, we concatenate the color values of each sample with the direction vector of where it was sampled from.
Patches are then projected to the dimensions of the transformer using an SO(2)-equivariant fully connected layer. 
We refer to the output of the projection as patch embeddings.
Following the Vision Transformer paper, we prepend an output token to the patch embeddings~\cite{VisionTransformer2021}.
This token is only learnable in the invariant dimensions of our input, in the equivariant dimensions, it is set to zero. 
Having the token be learnable in any of the equivariant dimensions would break equivariance. %
The patch embeddings are then fed into a standard SO(3)-equivariant VN-Transformer~\cite{VNTransformer2023}.
The output of the transformer at the output token serves as a representation for the full image.
This output is then projected from the transformer dimension to the latent dimension using two SO(2)-equivariant fully connected layers, which compute $\mu$ and $\sigma$ that, following the reparameterization trick~\cite{vae2014}, yield the latent code. %
We found that using an SO(2)-equivariant input and output projection with a VN-Transformer encoder produces the best results.

\noindent\textbf{Rotation-Equivariant Patching.}
We sample patches from the spherical image using our SO(2)-equivariant patching strategy, as shown in Figure~\ref{fig:model}.
Patches are sampled as vertical stripes on the sphere.
A rotation of the environment map around the up axis by a multiple of the stripe width
results in a permutation of the color values in the patches.
Since the transformer is permutation invariant, having only this without any positional encoding would make the model rotation invariant.
The patches become equivariant by adding a directional component in the form of direction vectors to each pixel by concatenating the color values with the direction vectors.
A rotation of the environment map around the up-axis by a multiple of the stripe width, while keeping the patches in place,
is equivalent to a joint rotation of both the patches and the environment map around the up-axis.
Both result in the same rotated patch embeddings. Thus, our patching strategy is SO(2)-equivariant.
In practice, we use 64 patches as a trade-off between memory usage and rotation-equivariance.
Combining the directional vectors with the color values, is proposed in a similar way as early fusion of non-spatial attributes in~\cite{VNTransformer2023}.
We use early fusion rather than late fusion since the direction vectors only provide directional information and do not hold any additional information.

Since we work with spherical data but sample pixels from an equirectangular projection of spherical images, we have to ensure that the sampling is regularly distributed on the sphere. 
We evenly sample the azimuth $\varphi$ and distribute the polar angle $\theta$ as:
\begin{equation}
    \theta = \arccos(x), \, x\in[-1,1]
\end{equation}
By handling spherical data directly instead of working with projected images, 
we do not have to account for the distortion that would otherwise be introduced by the 3D to 2D projection.
Our approach of combining direction vectors and color values enables the model to know where each pixel was sampled from, eliminating the need for distortion correction. %

\noindent\textbf{Rotation-Equivariant Variational Sampling.}
Since our latent space is in 3D, variational sampling has to also take place in 3D space.
The naive approach of sampling three 1D distributions results in an axis-aligned distribution, which is not rotation-equivariant.
To solve this issue, we sample from a spherical normal distribution that is defined by a 3D mean vector and a 1D variance.
This distribution is isotropic and thus rotation-equivariant.

\noindent\textbf{RENI++ decoder.}
We adopt the same decoder as proposed in RENI++~\cite{RENI++2023},
which introduces a rotation-equivariant conditional neural field autodecoder architecture that leverages attention for conditioning. This design is motivated by findings that attention-based conditioning of neural fields outperforms alternative methods of conditioning~\cite{AttentionNeuralField2022}.
The input to the RENI++ neural field $\mathcal{D}$ is a direction vector $\mathbf{d}$ and the 3D latent vector $\mathbf{Z}$.
The output is the color $\mathbf{c}$ at direction $\mathbf{d}$.
To achieve rotation-equivariance, $\mathbf{d}$ and $\mathbf{Z}$ are encoded so they are invariant to simultaneous rotations, so $\mathcal{D}(\mathbf{Rd},\mathbf{RZ})=\mathcal{D}(\mathbf{d},\mathbf{Z})$ with $\mathbf{R}\in\textit{SO}(3)$.
This makes the neural field equivariant to rotations of $\mathbf{Z}$ only. 
Rotating $\mathbf{Z}$ is equivalent to rotating the spherical signals such that $\mathcal{D}(\mathbf{d},\mathbf{RZ})=\mathcal{D}(\mathbf{R}^\top\mathbf{d},\mathbf{Z})$. 
The invariant encoding of $\mathbf{d}$ and $\mathbf{Z}$ to simultaneous rotations is achieved by encoding the direction $\mathbf{d}$ relative to the latent code: $\mathbf{d'}=\mathbf{Z}^\top\mathbf{d}$ and applying the Vector Neuron invariant layer~\cite{VectorNeurons2021} to the latent code: $\mathbf{Z'}=\textrm{VN-Inv}(\mathbf{Z})$.

\noindent\textbf{Pretraining.}
For rendering, the full dynamic range of natural light is essential.
To let our model learn this full range, our dataset must also consist of images that cover the entire dynamic range of natural light and not clipping the brightest parts of the image.
We use two datasets to train our model.
One is the RENI++ dataset, consisting of 1,694 HDR equirectangular images obtained under a CC0 1.0 Universal Public Domain Dedication license~\cite{RENI++2023}. 
It features a wide variety of outdoor scenes and illumination conditions.
These images have a resolution of $128\times 64$. This low resolution is sufficient for our purposes.
The RENI++ dataset is limited in size and further availability of 360° HDR outdoor illumination environments is limited.
However, 360° LDR data is readily available, for example in the form of Google street view data.
Although we cannot use LDR data directly, there exists a large body of work that focuses on reconstructing HDR data from LDR images~\cite{LANetHDR2021, lediff2025, HDRReconstruction2017, hdrsky2023}. %
Some of which even take 360° environment maps into account~\cite{LANetHDR2021,hdrsky2023}.
In addition to the RENI++ dataset, we also use the streetlearn dataset~\cite{streetlearn2019}, consisting of a large number of LDR 360° images captured by Google street view in Manhattan.
We convert 43,310 equirectangular 360° images from the Manhattan part of the streetlearn dataset to HDR using~\cite{LANetHDR2021}.
The converted streetlearn dataset is of lower quality than the RENI++ dataset since the HDR reconstruction process is imperfect and the dataset predominantly contains cityscapes, 
whereas we want good HDR reconstruction of varied outdoor scenes. %
We address this in our training curriculum by first pretraining our model on the much larger streetlearn dataset and then finetuning it on the RENI++ dataset.

\section{Experiments}
\label{sec:experiments}

\noindent\textbf{Losses.}
To better match how the human visual system reacts to luminance, we train our model in $\log$ space. 
This is common practice for models that deal with HDR data~\cite{LANetHDR2021, lediff2025, HDRReconstruction2017, hdrsky2023}.
With a loss in linear space, areas with high luminance would overpower important features in areas with low or medium luminance. 
A loss in $\log$ space alleviates this issue by spreading the loss approximately linearly across the perceived luminance range~\cite{HDRReconstruction2017}.

Additionally, there is scale ambiguity within our dataset, because the HDR images are captured at unknown exposure values (EV).
We only know the relative luminance between pixels, not the absolute luminance of each pixel.
Any image multiplied by a positive scale factor $k$, would still be valid. %

Our dataset contains high frequency details, which are especially prevalent at the boundary between sun and sky, where there is a large change in intensity in a small area.
To enable our model to capture more high frequency detail, we use a loss from depth prediction, the Mean Absolute Gradient Error (MAGE)~\cite{DepthPro2025}.
This loss improves high frequency detail by operating on the gradients of the image. 
High frequency details will lead to high gradients and thus the error in high frequency areas is more strongly penalized.
During training, we decode the full equirectangular images. Thus, we have to account for the irregular sampling by weighting with the $\sin$ of the polar angle of each pixel, $\sin(\theta_i)$:
\begin{equation}
    \mathcal{L}_{\textit{MAGE}} = \frac{1}{M}\sum_{j=1}^M\frac{1}{N_j}\sum_{i=1}^{N_j}\sin(\theta_i)\lvert \nabla_Sf(\mathbf{I})^j_i-\nabla_S\mathbf{I}_i^j\rvert,
\end{equation}
with $\nabla_S$ denoting the spatial derivative operator Scharr (S)~\cite{Scharr1997} and $M=2$  denoting the number of scales, used for the multi-scale derivative loss. 
The inputs to the loss are $\mathbf{I}$, the ground truth image in $\log$ space, and $f(\mathbf{I})$, the model output in $\log$ space.
Since this loss is applied in $\log$ space, it is scale invariant. 
In $\log$ space, Scharr removes the influence of any arbitrary scale factor $k$.

We also follow the approach of RENI++ and LANet~\cite{RENI++2023, LANetHDR2021} and use another technique from depth prediction~\cite{DepthMapPrediction2014, MegaDepth2018}, the scale invariant loss.
This loss calculates the relative error over the entire image, rather than on the gradients of the image:
\begin{equation}
    \mathcal{L}_{\textit{scale-inv}} = \frac{1}{N}\sum_{i=1}^N\left(R_i\right)^2-\frac{1}{N^2}\left(\sum_{i=1}^NR_i\right)^2.
\end{equation}

where $R_i=\sin(\theta_i)(f(\mathbf{I})_i-\mathbf{I}_i)$ at pixel $i$. This helps us learn a scale invariant representation of the images.
To encourage our model to produce accurate color representations, we again follow the approach of RENI++~\cite{RENI++2023} and use a cosine similarity loss.
Since this loss operates on the direction of RGB vectors, it is also scale invariant:
\begin{equation}
    \mathcal{L}_{\textit{cosine}}=1-\frac{1}{N}\sum_{i=1}^N\sin(\theta_i)\frac{f(\mathbf{I})_i \cdot \mathbf{I}_i}{\lVert f(\mathbf{I})_i \rVert \lVert \mathbf{I}_i\rVert}.
\end{equation}

We train our model using variational sampling and regularize the latent space to follow a standard normal distribution using Kullback Leibler divergence (KLD).
Since we use a 3D isotropic distribution for our variational sampling, we must also adjust the KLD loss term accordingly:
\begin{equation}
  \mathcal{L}_\textit{KLD} = \frac{1}{D}\sum_i^D-0.5 \cdot  (3 + 3\cdot\log(\sigma_i) - \lVert\mu_i\rVert^2 - 3\cdot\sigma_i),
  \label{eq:3d-kld}
\end{equation}
where $D$ is the latent dimension. Our full training loss is:
\begin{equation}
    \mathcal{L} = 0.5\cdot\mathcal{L}_\textit{MAGE}+\mathcal{L}_{\textit{scale-inv}}+\mathcal{L}_\textit{cosine}+0.01\cdot\mathcal{L}_\textit{KLD}.
    \label{eq:combined-loss}
\end{equation}

\begin{table*}[t]
\centering
\begin{tabularx}{\linewidth}{@{\extracolsep{6pt}}cXXXXXXXXXXXX}
\toprule & \multicolumn{4}{c}{RENI++ (optimization)} & \multicolumn{4}{c}{Ours (AE)} & \multicolumn{4}{c}{Ours (optimization)} \\
\cmidrule{2-5} \cmidrule{6-9} \cmidrule{10-13} 
$D$ 
& PSNR$\uparrow$ & SSIM$\uparrow$ & LPIPS$\downarrow$ & PSNR HDR$\uparrow$
& PSNR$\uparrow$ & SSIM$\uparrow$ & LPIPS$\downarrow$ & PSNR HDR$\uparrow$
& PSNR$\uparrow$ & SSIM$\uparrow$ & LPIPS$\downarrow$ & PSNR HDR$\uparrow$ \\
\midrule
27   & 18.02 & 0.39 & 0.62 &33.00& 18.78 & 0.46 & 0.62 &33.37& \textbf{20.33} & \textbf{0.51} & \textbf{0.61}&\textbf{33.92}\\
147  & 21.13 & 0.51 & \textbf{0.55} &34.30& 19.40 & 0.48 & 0.59 &33.59& \textbf{21.89} & \textbf{0.54} & \textbf{0.55}&\textbf{35.07}\\
300  & 22.10 & 0.55 & 0.52 &35.10& 19.47 & 0.48 & 0.59 &33.68& \textbf{22.68} & \textbf{0.57} & \textbf{0.48} &\textbf{35.59}\\

\bottomrule
\end{tabularx}
\caption{Comparison of reconstruction quality metrics between RENI++ and our model (autoencoder pass and direct latent code optimization), across varying latent dimensions $D$.  PSNR, SSIM and LPIPS are in LDR tone-mapped space, PSNR HDR is in linear HDR space.}
\vspace{-0.05in}
\label{tab:reconstruction}
\end{table*}

\noindent\textbf{Quality Metrics.}
We report in Table~\ref{tab:reconstruction}, the PSNR, SSIM and LPIPS scores in LDR tone-mapped space, as well as PSNR in linear HDR space, all averaged across the test set. 
RENI++ results are taken from their paper, as we found them reproducible. 
For our model, we report metrics for both the full autoencoder pass and decoder-only latent optimization, the latter being more comparable to RENI++ which supports only latent optimization.
Following RENI++, we also optimize a per-image scale factor $k$ during latent optimization. This factor scales the decoder output and is not considered in the full autoencoder pass.
Especially in the low-dimensional case of $D=27$, our model strongly outperforms RENI++. 
We include a qualitative comparison of the three approaches in the supplementary material.

\begin{figure}
  \centering
   \includegraphics[width=\linewidth]{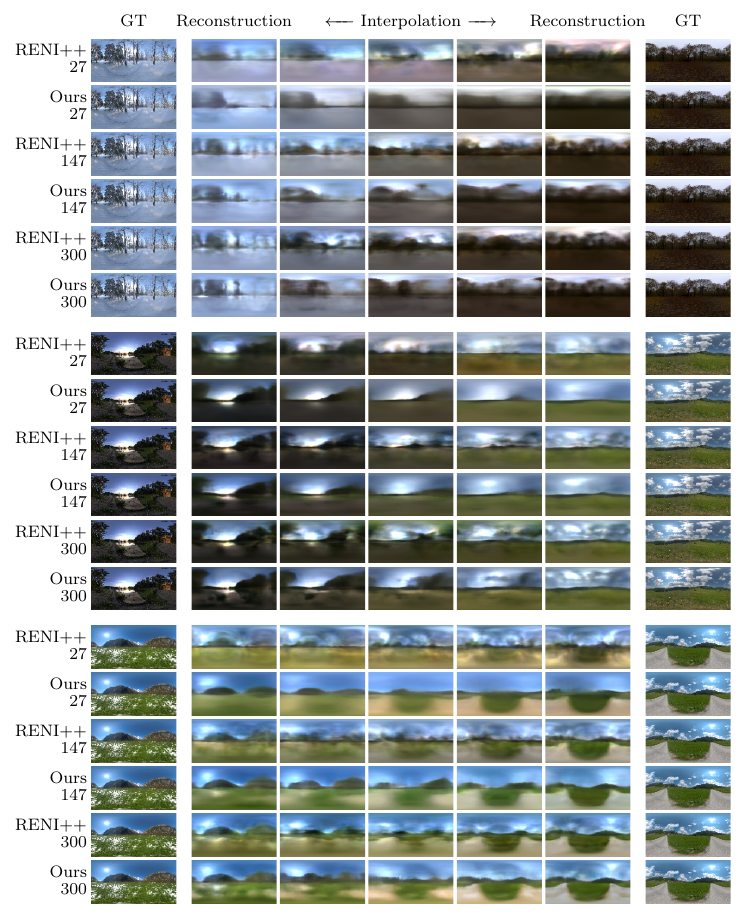}
   \caption{Interpolations using our model (with direct latent optimization) and RENI++ with different latent sizes. Three image pairs are interpolated: snow-forest, lake-field, mountain-road.}
   \label{fig:interpolation}
\end{figure}

\noindent\textbf{Interpolation.}
A well-behaved latent space enables smooth interpolations between latent codes.
Each step of the interpolation should correspond to a realistic and semantically meaningful illumination environment.
For example, interpolating between a sunrise and a midday illumination environment 
should result in the sun moving smoothly across the sky and the color temperature changing gradually from daylight to sunset without introducing artifacts.

Figure~\ref{fig:interpolation} shows interpolations of RENI++ and our model for the same image pairs. 
Our model meaningfully interpolates between the start and target image, whereas RENI++ introduces artifacts in the interpolation steps. 
In the snow-forest example, our model shows a smooth color change, while RENI++ introduces a sun in the middle of the interpolation, that is not present in either the source or the target image.
Our model can also smoothly interpolate between sunrise and midday illumination environments, as seen in the lake-field example. The yellow reflection in the lake smoothly disappears and the scene transitions to a more bluish midday illumination environment.
In the mountain-road interpolation, RENI++ introduces artifacts and noise in the sky during interpolation, whereas our model does not. This is especially visible with $D=27$.

\begin{figure*}
  \centering
   \includegraphics[width=0.8\linewidth]{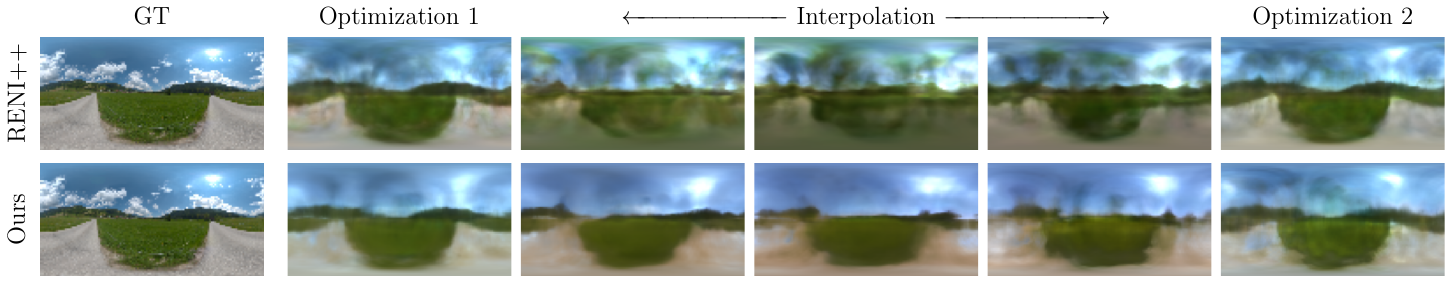}
   \caption{Interpolating between two optimized latent codes. With our model, the output images remain consistent during interpolation, suggesting that the latent space is unique.
With RENI++ the output images diverge from the optimized images during interpolation, suggesting that the latent space is not unique.}
   \label{fig:uniqueness}
\end{figure*}

\begin{table}[t]
\centering
\begin{tabularx}{\linewidth}{@{\extracolsep{6pt}}ccccc}
\toprule & \multicolumn{2}{X}{Uniqueness$\downarrow$} & \multicolumn{2}{X}{Reconstruction Consistency$\uparrow$}  \\
\cmidrule{2-3}\cmidrule{4-5}
$D$ & RENI++ & Ours & RENI++ & Ours  \\
\midrule
27   & 1.46& \textbf{0.04}&0.17 & \textbf{0.50}\\
147  & 1.11& \textbf{0.43}&0.12 & \textbf{0.30}\\
300  & 1.03& \textbf{0.57}&0.07 & \textbf{0.23}\\
\bottomrule
\end{tabularx}
\caption{Uniqueness is the mean squared error (MSE) between an image produced by an optimized latent code and an image produced by the interpolation midpoint of two latent codes optimized to the same image.
Reconstruction Consistency are the Spearman correlation coefficients between the MSE of random latent pairs and the MSE between their corresponding image reconstructions. 
}
\vspace{-0.1in}
\label{tab:uniqueness_correlation}
\end{table}

\begin{table}[t]
\centering
\begin{tabular}{cccc}
\toprule
Dataset size & RENI++ & Ours (AE) & Ours (optimized) \\
\midrule
1500   & 20.11 & 16.00 & \textbf{20.99} \\
43260  & 17.14 & 16.77 &  \textbf{19.90} \\

\bottomrule
\end{tabular}
\caption{Mean PSNR of our model and the RENI++ model trained on a large and a small subset of the HDR converted streetlearn dataset, evaluated on the same test set from the converted streetlearn dataset.}
\vspace{-0.1in}
\label{tab:dataset_size}
\end{table}

\begin{figure}
 \centering
\includegraphics[width=0.99\linewidth]{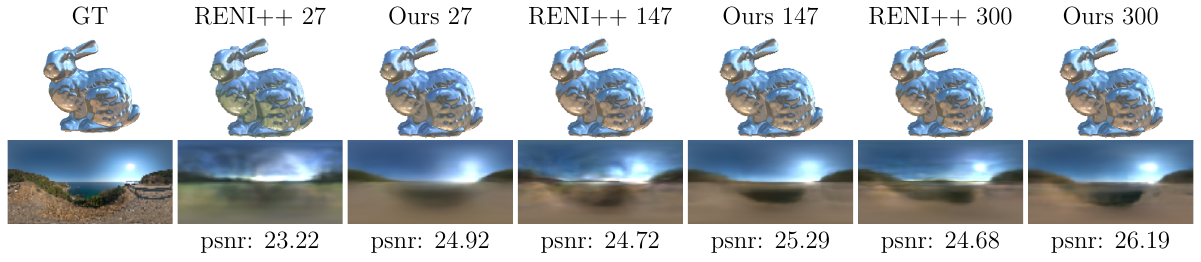}
\caption{Application: the latent space of our model is more amenable to inverse rendering optimization, outperforming the RENI++ baseline.}
\label{fig:inverse_rendering}
\end{figure}

\noindent\textbf{Uniqueness.}
One goal of our model is uniqueness of the latent space.
To test this, we optimize two random latent codes sampled from a normal distribution $\mathcal{N}(0,1)$ to fit the same image.
If the latent space is unique, no other images should be represented between the two optimized latent codes. %
We measure model uniqueness by interpolating between two optimized latent codes, as shown in Figure~\ref{fig:uniqueness}.  
For unique models, the output images do not diverge during interpolation. 
Since the output images barely diverge for our model and diverge significantly for RENI++, we can conclude that our model is more unique than RENI++.
To objectively measure uniqueness, we compute the MSE between the image produced by the first optimized latent code and the image produced by the latent interpolation midpoint.
Higher error indicates that the model is less unique.
Table~\ref{tab:uniqueness_correlation} shows this metric for our model and the RENI++ model over various latent dimensions $D$.
We include further examples of the interpolation between two optimized latent codes in the supplementary material.

The RENI++ autodecoder architecture hinders uniqueness in the model. Two similar images may be initialized with vastly different latent codes, while our model encodes two similar images to similar latent codes by using a variational autoencoder architecture.
With the small dataset RENI++ uses for training, non-uniqueness is not a significant issue.
However, with a larger dataset, there are more similar images that are then initialized with different latent codes, decreasing the uniqueness of the model even further.
This also leads to a decrease in RENI++ performance as dataset size increases, as seen in Table~\ref{tab:dataset_size}.
While RENI++ performance degrades with an increased dataset size, our model's performance improves with an autoencoder pass and does not degrade as much with optimization.

We also measure the reconstruction consistency of our model (see Table~\ref{tab:uniqueness_correlation}). This metric calculates the correlation coefficients between the error of random latent pairs and the error between their corresponding image reconstructions.
Following the idea of the latent-data distance constraint by Tran \etal \cite{distgan2018}, it tells us how well latent distance matches perceived distance and is an indicator for latent space uniqueness. %
A higher correlation indicates more uniqueness because a low latent error combined with a low reconstruction error (\ie, a more unique model) results in a higher correlation. In contrast, a high latent error combined with a low reconstruction error (\ie, a non-unique model) results in a lower correlation.
The improved reconstruction quality and more unique latent space of our model enable improvements in downstream applications such as inverse rendering optimization, as shown in Figure~\ref{fig:inverse_rendering}.

\begin{table}[t]
\centering
\begin{tabular}{cccc}
\toprule
Component & 27 & 147 & 300 \\
\midrule
Ours & \textbf{18.78} & \textbf{19.40} & \textbf{19.47} \\
- scale-inv \& MAGE loss & 17.32 & 18.47 & 18.80 \\
- pretraining on streetlearn & 17.37 & 18.07 & 17.88 \\ %
- SO(2) linear projection  & 16.89 & 17.49 & 17.30 \\

\bottomrule
\end{tabular}
\caption{Mean PSNR on various ablations of our model}
\vspace{-0.05in}
\label{tab:various_ablations}
\end{table}

\begin{table}[t]
\centering
\begin{tabular}{cccc}
\toprule
$D$ & SO(2) projections & Full SO(2) & Full VN \\
\midrule
27  & \textbf{18.78} & 17.89 & 17.74 \\ 
147 & \textbf{19.40} & 18.18 & 18.80 \\
300 & \textbf{19.47} & 17.93 & 18.60 \\

\bottomrule
\end{tabular}
\caption{Mean PSNR comparing models only using SO(2) linear layers in the projection layers, using SO(2) layers in the entire encoder including the transformer and using VN-layers in the entire encoder.}
\vspace{-0.1in}
\label{tab:so2_ablations}
\end{table}

\noindent\textbf{Ablations.}
To show that our SO(2)-equivariant extension to Vector Neurons improves upon the default SO(3)-equivariant Vector Neurons, we do an ablation study.
We compare the performance of Vector Neurons with that of our SO(2)-equivariant extension in the projection layers and the full encoder across a range of latent dimensions $D$ using the mean PSNR.
For the SO(2)-equivariant transformer, we have to account for the SO(2)-equivariant layers using the invariant axes as additional learnable parameters, 
by decreasing the inner dimension of the SO(2)-equivariant transformer accordingly.
Table~\ref{tab:so2_ablations} shows the results of this evaluation.
The best results are achieved when the SO(2)-equivariant fully connected layers are used only as projection layers.
Thus, to benefit from reducing the equivariance guarantee from SO(3) to SO(2),
it is not necessary to reduce equivariance for the entire model.
Reducing the equivariance by using SO(2)-equivariant fully connected layers as projection layers is sufficient.

We also test the losses used over MSE, as well as our pretraining method that uses the converted streetlearn dataset.
Table~\ref{tab:various_ablations} shows the results for the full autoencoder pass, %
comparing mean LDR PSNR over various ablations and latent dimensions.
We include a qualitative comparison of the ablations in the supplementary material.

\section{Conclusion}
\label{sec:conclusion}

Our model produces realistic HDR environment maps with a well-behaved latent space that enables smooth interpolation.
We design a rotation-equivariant vision transformer (VN-ViT) that operates on 360° panoramic images and follows the Vector Neuron principles. We introduce an SO(2)-equivariant extension to Vector Neurons and use it to reduce the equivariance of the VN-ViT from SO(3) to SO(2), resulting in an increased performance.
To further enhance our model's performance, we adapt losses from depth prediction.
Our model outperforms RENI++ in terms of reconstruction quality and latent-space structure.
It provides a more unique latent representation and scales effectively to larger datasets, whereas RENI++ performance degrades, our model continues to improve. 
In conclusion, our work advances the modeling of natural illumination, and we believe this contribution will benefit downstream tasks such as inverse rendering, relighting and other vision problems that rely on accurate illumination priors.

\noindent\textbf{Acknowledgments.}
The authors gratefully acknowledge the scientific support and HPC resources provided by the Erlangen National High Performance Computing Center (NHR@FAU) of the Friedrich-Alexander-Universität Erlangen-Nürnberg (FAU). The hardware is funded by the German Research Foundation (DFG).
James Gardner was supported by the EPSRC Centre for Doctoral Training in Intelligent Games \& Games Intelligence (IGGI) (EP/S022325/1).

\newpage
{
    \small
    \bibliographystyle{ieeenat_fullname}
    \bibliography{main}
}

\clearpage
\setcounter{page}{1}
\maketitlesupplementary

\setcounter{section}{0}
\renewcommand{\thesection}{\Alph{section}}
\renewcommand{\thesubsection}{\Alph{section}.\arabic{subsection}}

\section{Proof of SO(2)-rotation-equivariance}

In section~\ref{sec:methods}, we introduced an SO(2)-rotation-equivariant fully connected layer.
Here we prove the SO(2)-rotation-equivariance of this layer.

Let $R\in\textit{SO}(2)$ be a $2\times 2$ rotation matrix and:
\[R'=\begin{pmatrix}
R & 0\\
0 & \mathbf{I}_{c_\textit{inv}}
\end{pmatrix} \; ,\]
where $\mathbf{I}_{c_\textit{inv}}$ is the identity matrix of size $c_\textit{inv}\times c_\textit{inv}$.
From the definition of $R'$, it follows that

\begin{align}
    R'\mathbf{X}=\left[R\mathbf{X}_\textit{eq},\mathbf{X}_\textit{inv}\right]\;.
\end{align}

The intermediate values $\mathbf{T}_\textit{inv}$ and $\mathbf{T}'_\textit{inv}$ are invariant to the rotation:

\begin{align}
    \mathbf{T}_{\textit{inv}} &= \left[\mathbf{X}_\textit{inv},\mathbf{1}\right] \\
    \mathbf{T}'_{\textit{inv}} &= \left[ \mathbf{X}_\textit{inv},\lVert R\mathbf{X}_\textit{eq}\rVert\right] = \left[ \mathbf{X}_\textit{inv},\lVert \mathbf{X}_\textit{eq}\rVert\right]
\end{align}

Therefore, $Y_\textit{inv}$ is also invariant to the rotation:

\begin{align}
  \mathbf{Y}_{\textit{inv},o,v} &=\sum_{i=1}^{d_\textit{in}}\sum_{k=1}^{c_\textit{inv}+1}\mathbf{W}_{\textit{inv},o,v,i,k}\cdot \mathbf{T}'_{\textit{inv},i,k}+\mathbf{B}_{\textit{inv},o,v}
\end{align}

For the $Y_\textit{eq}$ the rotation commutes:

\begin{align}
    \mathbf{Y}'_{\textit{eq},o,v} &= \sum_{i=1}^{d_\textit{in}}\sum_{k=1}^{c_\textit{inv}+1} \mathbf{W}_{\textit{eq},o,k,i}\cdot \mathbf{T}_{\textit{inv},i,k}\cdot (R\mathbf{X}_{\textit{eq}})_{i,v} \\
    &= \sum_{i=1}^{d_\textit{in}}\sum_{k=1}^{c_\textit{inv}+1} \mathbf{W}_{\textit{eq},o,k,i}\cdot \mathbf{T}_{\textit{inv},i,k}\cdot (\sum_{t=1}^2 R_{v,t}\mathbf{X}_{\textit{eq},i,t})\\
    &= \sum_{t=1}^2 R_{v,t}\sum_{i=1}^{d_\textit{in}}\sum_{k=1}^{c_\textit{inv}+1} \mathbf{W}_{\textit{eq},o,k,i}\cdot \mathbf{T}_{\textit{inv},i,k}\cdot \mathbf{X}_{\textit{eq},i,t}\\
    &= \sum_{t=1}^2 R_{v,t}\mathbf{Y}_{\textit{eq},o,t}\\
    &= (R\mathbf{Y}_{\textit{eq}})_{o,v}
\end{align}

We have shown that: \[f\left(\left[R\mathbf{X}_\textit{eq},\mathbf{X}_\textit{inv}\right], \mathbf{W}_\textit{eq},\mathbf{W}_\textit{inv},\mathbf{B}_\textit{inv}\right) = \left[ R\mathbf{Y}_\textit{eq},\mathbf{Y}_\textit{inv}\right]\]
\[f\left(R'\mathbf{X}, \mathbf{W}_\textit{eq},\mathbf{W}_\textit{inv},\mathbf{B}_\textit{inv}\right) = R'f\left(\mathbf{X}, \mathbf{W}_\textit{eq},\mathbf{W}_\textit{inv},\mathbf{B}_\textit{inv}\right)\]
The rotation matrix commutes with this SO(2)-equivariant fully connected layer and thus,
we can conclude that our SO(2)-equivariant fully-connected layer has the desired SO(2)-equivariance and invariance properties.

\section{SO(2)-rotation-equivariant transformer}
In section~\ref{sec:experiments}, we use an SO(2)-equivariant transformer for ablations.
This transformer is built by replacing every Vector Neuron component of the VN-Transformer~\cite{VNTransformer2023} with SO(2) equivariant layers. 
The VN-Linear layers are replaced with our SO(2)-equivariant fully connected layer, and the $\text{VN-ReLU}$ layers, 
are replaced with our $\text{SO(2)-ReLU}$ version, in which, the Vector Neuron version is applied in the equivariant dimensions $\mathbf{X}_\textit{eq}$
and the standard version in the invariant dimensions $\mathbf{X}_\textit{inv}$:

\begin{equation}
  \textrm{SO(2)-ReLU}(\left[ \mathbf{X}_\textit{eq}, \textbf{X}_\textit{inv}\right]) = \left[ \textrm{VN-ReLU}(\mathbf{X}_\textit{eq}), \textrm{ReLU}(\textbf{X}_\textit{inv})\right]
  \label{eq:sox-relu}
\end{equation}

\section{Additional Results}

\begin{figure*}
  \centering
   \includegraphics[height=0.95\textheight]{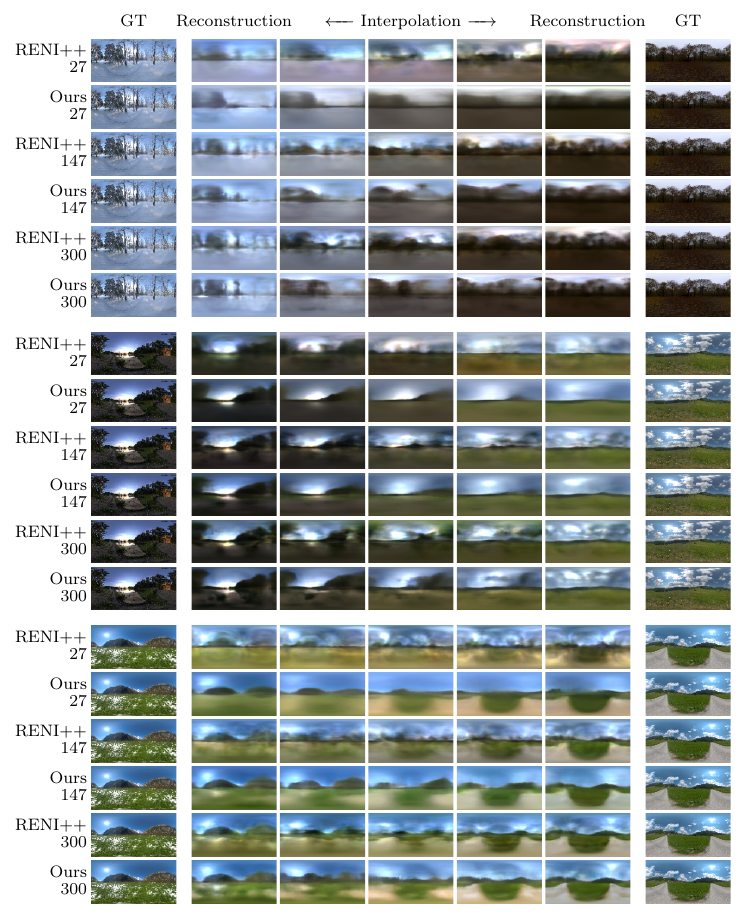}
\caption{Interpolations using our model (with direct latent optimization) and RENI++ with different latent sizes. Three image pairs are interpolated: snow-forest, lake-field, mountain-road.}
   \label{fig:interpolation_optimized}
\end{figure*}

\begin{figure*}
  \centering
   \includegraphics[height=0.95\textheight]{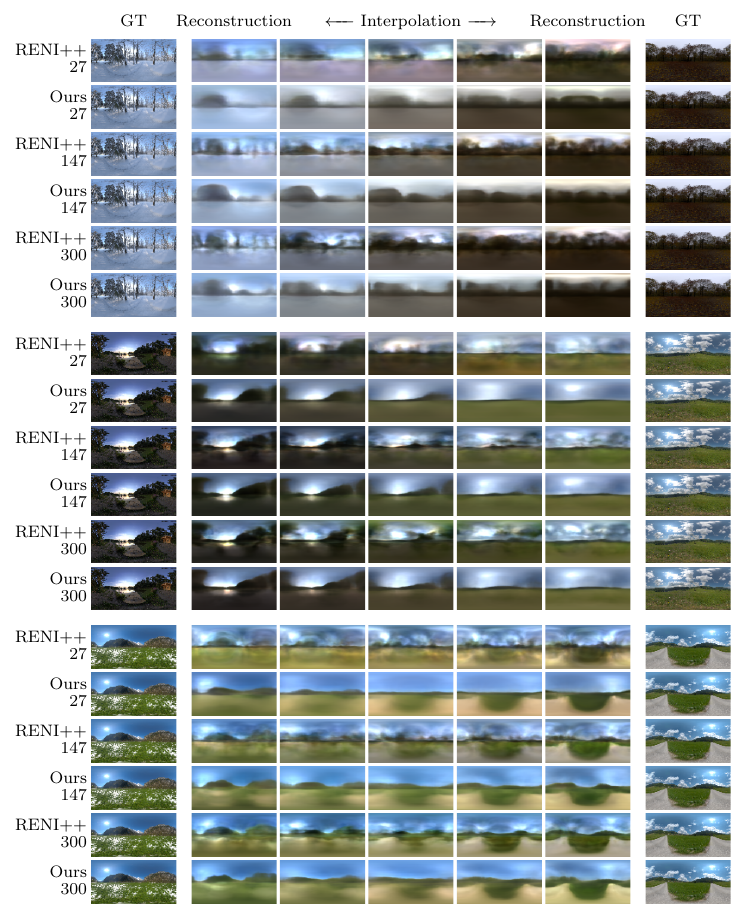}
   \caption{Interpolations using our model (autoencoder pass) and RENI++ with different latent sizes. Three image pairs are interpolated: snow-forest, lake-field, mountain-road.}
   \label{fig:interpolation_ae}
\end{figure*}

\begin{figure*}
  \centering
   \includegraphics[height=0.95\textheight]{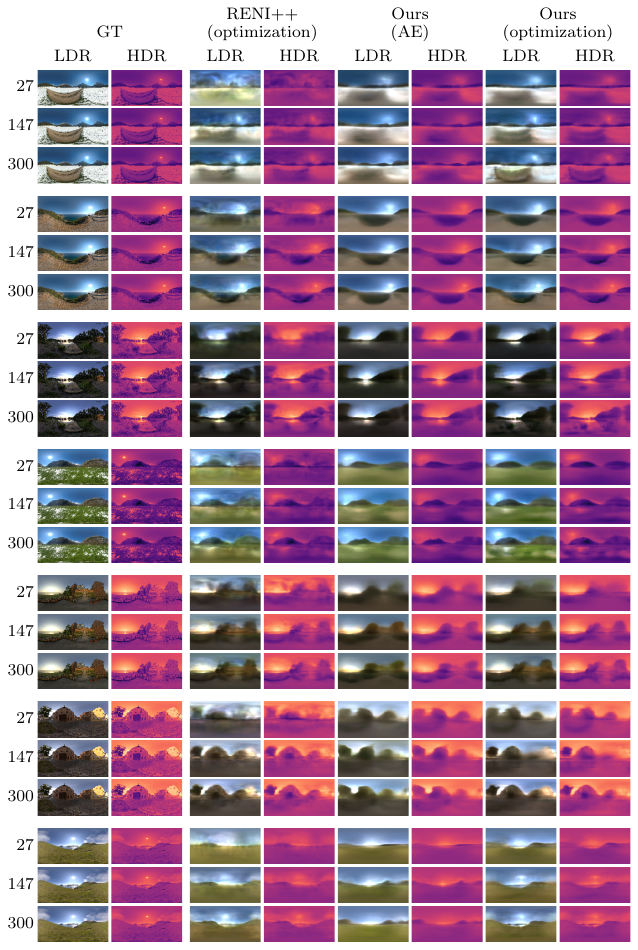}
   \caption{Corresponding figure to Table~\ref{tab:reconstruction} comparing RENI++ and our model (autoencoder pass and direct latent code optimization), across varying latent dimensions $D$ and target images, with reconstructions shown in LDR tone-mapped space and in log HDR space as heatmaps}
   \label{fig:reconstruction_quality}
\end{figure*}

\begin{figure*}
  \centering
   \includegraphics[width=\textwidth]{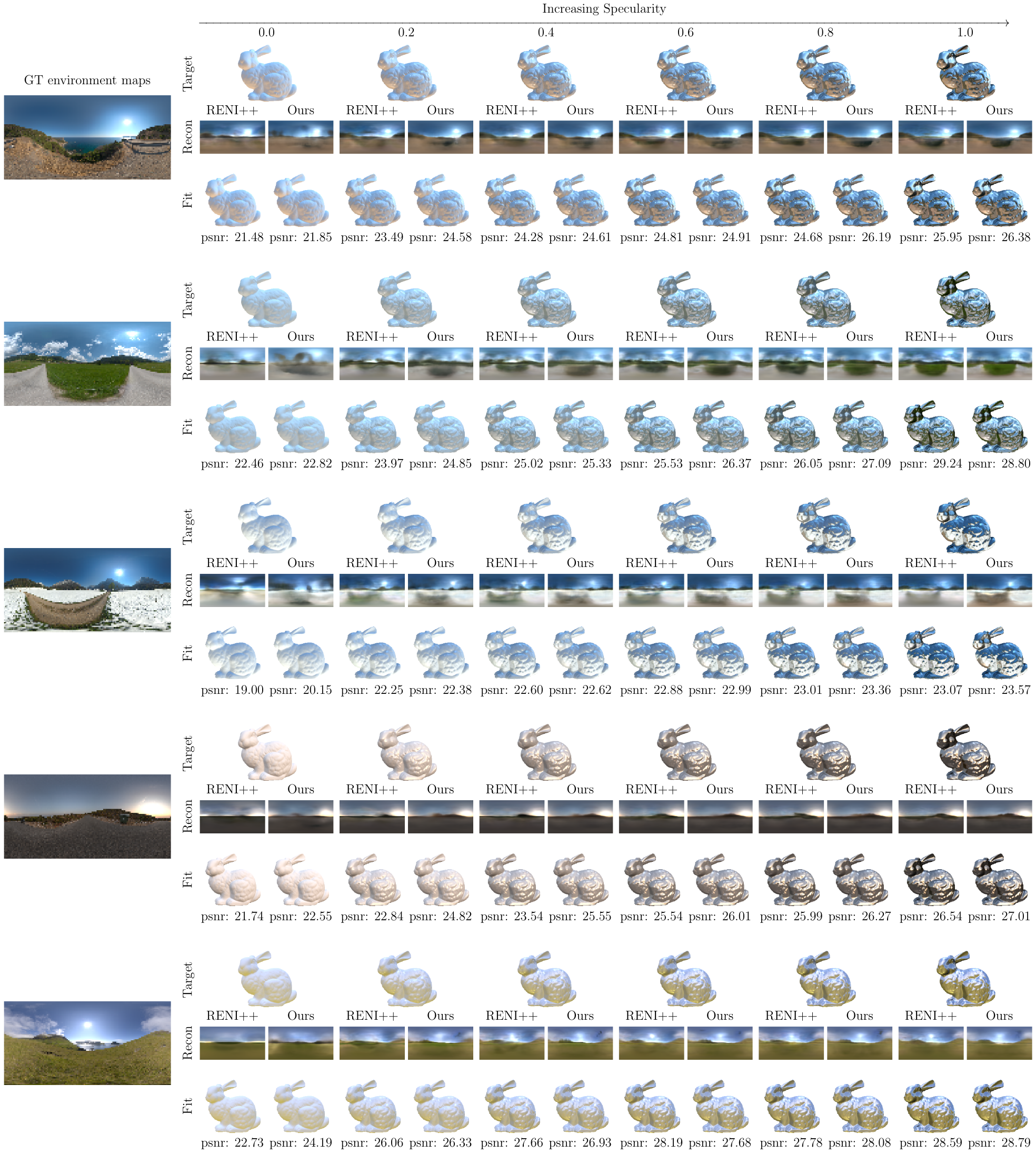}
   \caption{Further results of inverse rendering optimization using Blinn-Phong shading. We compare our model to RENI++, increasing the specular Blinn-Phong term $K_s$ from $0$ to $1$. Both RENI++ and our model have a latent dimensionality of $D=300$. Our model generally outperforms RENI++}
   \label{fig:suppl_inverse}
\end{figure*}

\begin{figure}
  \centering
   \includegraphics[width=\linewidth]{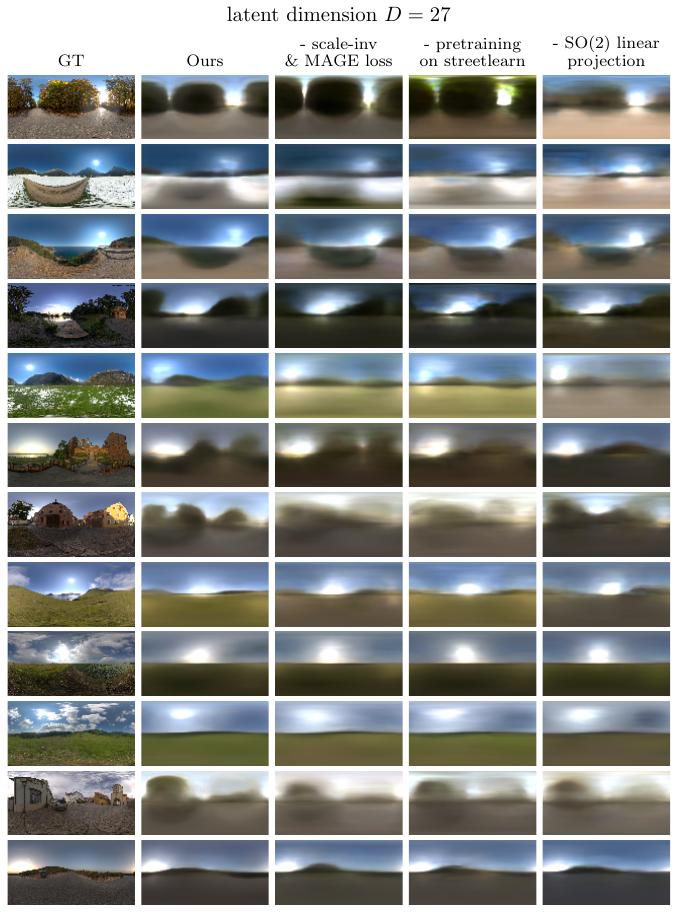}
   \caption{Corresponding figure to Table~\ref{tab:various_ablations} comparing various ablations of our model, here with latent dimension $D=27$.}
   \label{fig:ablations_27}
\end{figure}

\begin{figure}
  \centering
   \includegraphics[width=\linewidth]{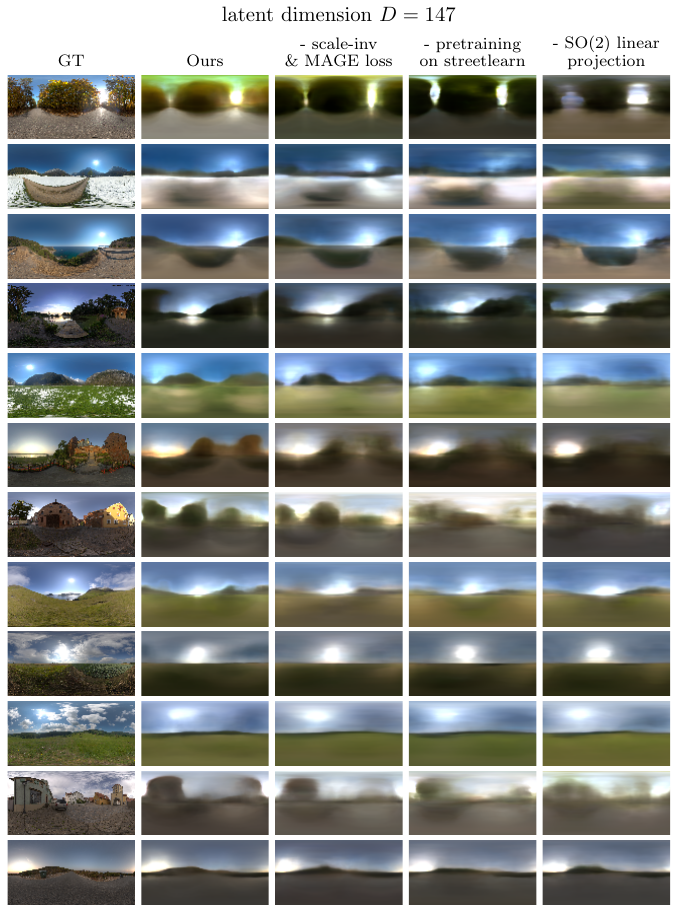}
   \caption{Corresponding figure to Table~\ref{tab:various_ablations} comparing various ablations of our model, here with latent dimension $D=147$.}
   \label{fig:ablations_147}
\end{figure}

\begin{figure*}
  \centering
   \includegraphics[height=0.95\textheight]{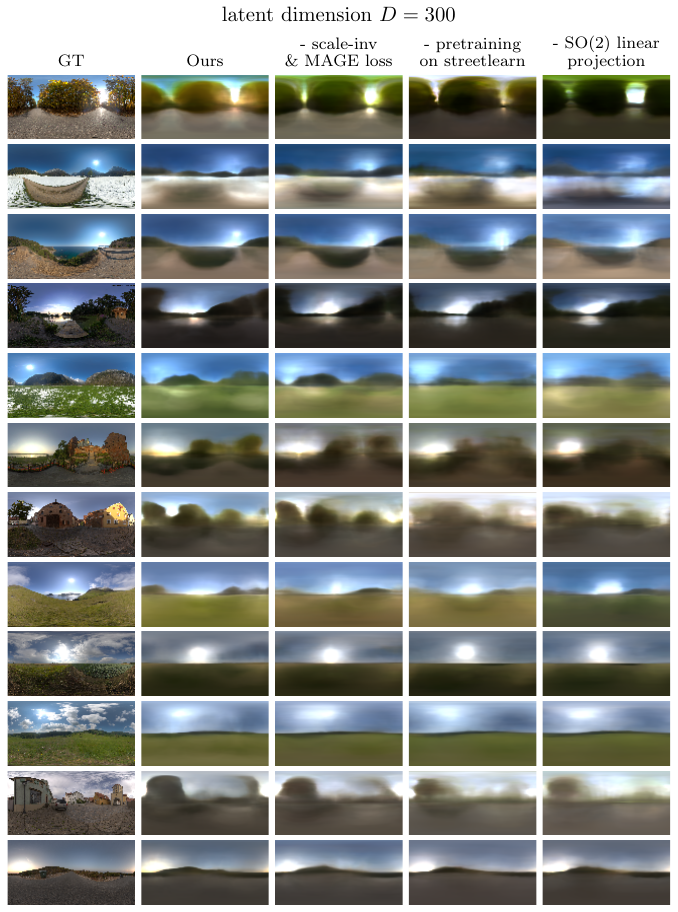}
   \caption{Corresponding figure to Table~\ref{tab:various_ablations} comparing various ablations of our model, here with latent dimension $D=300$.}
   \label{fig:ablations_300}
\end{figure*}

We present additional qualitative results for our model. Figures~\ref{fig:interpolation_optimized} and~\ref{fig:interpolation_ae} demonstrate interpolations using latent optimization and the autoencoder pass for the reconstructions. 
Figure~\ref{fig:reconstruction_quality} shows a comparison of the quality of reconstructions between RENI++ and our model. For our model, we present reconstructions obtained using the autoencoder pass and direct latent optimization. This figure corresponds to Table~\ref{tab:reconstruction}.
We also include an application of our model in Figure~\ref{fig:suppl_inverse}, which shows an inverse rendering optimization over an increasing Blinn-Phong specular term $K_s$ from $0$ to $1$.
Figures~\ref{fig:ablations_27},~\ref{fig:ablations_147} and~\ref{fig:ablations_300} show image reconstructions using the autoencoder pass for ablations of our model, corresponding to Table~\ref{tab:various_ablations}.
Figure~\ref{fig:uniquenes_large} includes additional examples following Figure~\ref{fig:uniqueness}, showing interpolations between latent codes optimized to the same GT image.

\begin{figure*}
  \centering
   \includegraphics[width=\textwidth]{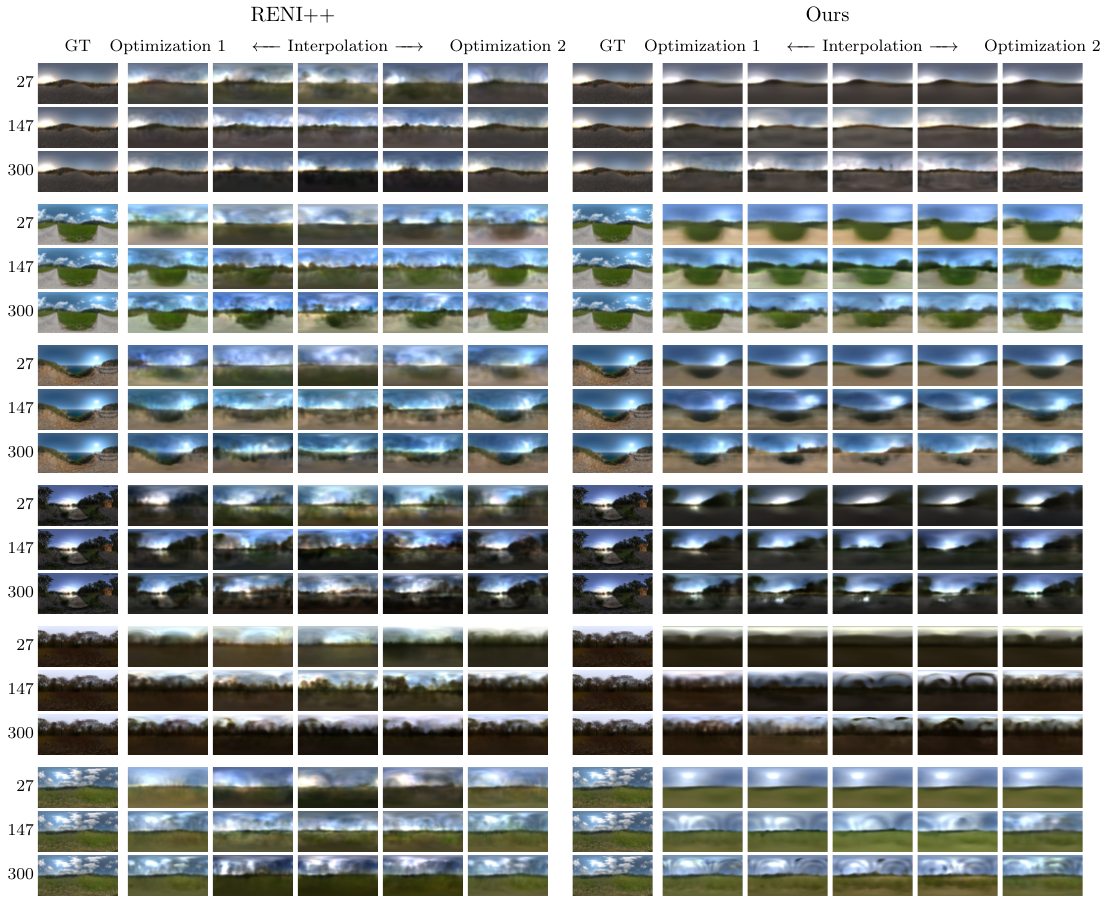}
   \caption{More examples following Figure~\ref{fig:uniqueness}. Comparing the uniqueness of RENI++ and our model across varying latent dimensions $D$ by interpolating between two latent codes optimized to the same target image.
   With our model, the output images remain relatively consistent during interpolation, suggesting that the latent space is unique.
   With RENI++ the output images diverge heavily from the optimized images during interpolation. There are color, luminance and structural shifts visible in the interpolation for RENI++, suggesting that the latent space is less unique.}
   \label{fig:uniquenes_large}
\end{figure*}

\end{document}